\documentclass[10pt,journal,compsoc]{IEEEtran}

\ifCLASSOPTIONcompsoc
  \usepackage[nocompress]{cite}
\else
  \usepackage{cite}
\fi

\usepackage{subcaption}
\usepackage{graphicx}
\usepackage{amsmath}
\usepackage{amssymb}
\usepackage{booktabs}
\usepackage{multirow}
\usepackage{siunitx}
\usepackage{xcolor}

\usepackage{dblfloatfix}

\begin{document}
%
\title{Fed-IGTD: A federated learning-based framework for tabular-to-image encoding}

\author{David Chushig-Muzo,
        Eva Milara,
        Francisco J. Lara-Abelenda,
        Luis Carlos Estrada-Petrocelli,
        Diego H. Peluffo-Ordóñez
\IEEEcompsocitemizethanks{
\IEEEcompsocthanksitem David Chushig-Muzo, and Eva Milara are with the Department of Signal Theory and Communications, Telematics and Computing Systems, Rey Juan Carlos University (URJC), Madrid, Spain. E-mail: \{david.chushig, eva.milara\}@urjc.es
\IEEEcompsocthanksitem Francisco J. Lara-Abelenda is with the Faculty of Experimental Sciences, Universidad Francisco de Vitoria, Madrid, Spain. E-mail: francisco.lara@ufv.es
\IEEEcompsocthanksitem Luis Carlos Estrada-Petrocelli is with Faculty of Engineering, Universidad Latina de Panamá (ULATINA), Panama City, Panama. E-mail: lestrada@ulatina.edu.pa
\IEEEcompsocthanksitem Diego H. Peluffo-Ordóñez is with the School of Mathematical  and Computational Sciences, Yachay Tech University, Ecuador. E-mail: dpeluffo@yachaytech.edu.ec
}
}


\IEEEtitleabstractindextext{%
\begin{abstract}
Tabular-to-image encoding methods, which transform tabular data into image-like representations, have shown promising results in predictive tasks. Despite their benefits, these methods follow a centralized paradigm, in which the data are collected and models are trained on a central device.  This raises privacy and security concerns, particularly the risk of data leakage, unauthorized data collection and access to users’ private information. Federated Learning (FL) provides a decentralized framework in which multiple clients jointly optimize a shared model without directly sharing training data. We introduce Fed-IGTD, a horizontal FL-based approach in which clients share a common feature space and upload only local statistics, from which the server reconstructs the global correlation matrix and obtains a feature-to-pixel mapping to train a convolutional neural network via Federated Averaging (FedAvg). We studied two differential privacy (DP) mechanisms: \textit{(i)} a one-shot Gaussian mechanism protecting the correlation-aggregation step; and \textit{(ii)} a client-level DP-FedAvg variant protecting the training with privacy loss composed via a R\'enyi-DP accountant. We evaluated Fed-IGTD on five datasets under a systematic sweep of non-IID severity and privacy budgets. Privacy protection applied to the mapping-construction step has little impact on predictive performance. By contrast, DP-FedAvg substantially degrades utility at moderate privacy budgets with a small client cohort ($K=5$). Ablation study indicates that this degradation is driven mainly by the injected noise rather than by uniform averaging or update clipping. Macro-F1 exhibits sensitivity to performance degradation than accuracy under non-IID heterogeneity and class imbalance. These findings have direct implications for cross-silo federated deployments characterized by a limited number of participating institutions and heterogeneous local data distributions. These results show the potential of tabular-to-image encoding methods for supporting privacy-preserving learning in decentralized settings.
\end{abstract}

\begin{IEEEkeywords}
Tabular-to-image methods, spatial encoding methods, synthetic images, IGTD, Tabular2image, federated learning
\end{IEEEkeywords}}

\maketitle

\IEEEdisplaynontitleabstractindextext

\IEEEpeerreviewmaketitle

\ifCLASSOPTIONcompsoc
\IEEEraisesectionheading{\section{Introduction}\label{sec:introduction}}
\else
\section{Background and motivation}
\label{sec:introduction}
\fi

\IEEEPARstart{T}{abular} data is ubiquitous across domains such as healthcare, finance, and bioinformatics~\cite{jiang2026representation}. Ensemble tree-based models have been extensively used for predictive tasks using tabular data because of their high performance~\cite{somvanshi2026survey}. The adoption of deep learning architectures has been challenging because these models are not inherently designed to handle the characteristics of tabular data. Unlike images, which exhibit an inherent spatial structure where neighboring pixels are related and convolutional filters can capture patterns, tabular data lack inherent spatial organization~\cite{wang2026robust}. 

Recent studies in representation learning for tabular data have led to significant progress~\cite{jiang2026representation}. In particular, transforming tabular data into synthetic images has enabled the application of image-based models to structured data~\cite{liu2025comprehensive}. Several tabular-to-image encoding methods have been proposed, including DeepInsight~\cite{sharma2019deepinsight}, REFINED~\cite{bazgir2020representation}, TINTO~\cite{castillo2023tinto}, IGTD~\cite{zhu2021converting}, TabSOM~\cite{chushig2026tabsom}. These methods perform a feature-to-pixel mapping such that features which present similarity in the input data are placed close together spatially. Obtaining this mapping requires a global view of inter-feature relationships across the dataset. This centralized paradigm is particularly challenging in high-stakes domains, where tabular data are both highly sensitive and distributed, such as hospitals holding patient records, financial institutions maintaining transaction histories, or geographically distributed sensor networks. In these settings, regulatory and privacy constraints often prohibit the centralization of raw data, posing significant challenges for training models that require data to be collected and processed in a single location.

Federated Learning (FL) is a novel paradigm to address these challenges, enabling distributed joint training using decentralized data across heterogeneous devices~\cite{mcmahan2017communication}. Instead of transmitting data from devices to the server, each device uses its local data to independently train a model and only shares model parameters with the server. These local parameters are then aggregated by the server to update the global model~\cite{chen2024federated}. FL keeps sensitive information on local devices, addressing privacy, and regulatory concerns, which are crucial in high-stakes domains such as healthcare, Internet of Things, finance among others~\cite{chen2024federated, shenoy2025exploring}. Although FL has shown excellent results, its application within tabular-to-image methods has been partially explored.

In this paper, we propose a horizontal FL-based tabular-to-image framework named Fed-IGTD. It reconstructs the global IGTD correlation matrix from local client statistics, without centralizing raw data. The server aggregates these statistics to compute the correlation matrix, then runs IGTD's optimization to derive a shared feature-to-pixel mapping, which are broadcast to all clients. Each client converts its local data into images using this shared mapping, and a Convolutional Neural Network (CNN) is trained across clients via Federated Averaging (FedAvg). We employ two Differential Privacy (DP) mechanisms: \textit{(i)} a one-shot Gaussian mechanism protecting the correlation-aggregation step; and \textit{(i)} a client-level DP-FedAvg variant protecting the full training loop, with privacy loss composed via a R\'enyi-DP accountant. 

The main contributions of this paper are the following:

\begin{itemize}

    \item A novel horizontal FL adaptation of IGTD in which the feature-to-pixel mapping is derived from a reconstructed global correlation matrix, obtained via additive aggregation of per-client statistics rather than access to raw data.

    \item A one-shot Gaussian mechanism that clips and perturbs client-side statistics before aggregation, providing a formal DP guarantee for the correlation-aggregation stage in the IGTD algorithm. 

    \item A client-level DP-FedAvg to protect the iterative CNN training process, with per-round clipping and Gaussian perturbation of client updates and privacy loss composed across training rounds using a R\'enyi DP accountant. 

    \item An evaluation of Fed-IGTD across five datasets spanning healthcare, finance, chemistry/agriculture, signal processing, and radar/physics. The evaluation considers controlled variations in non-IID severity and privacy budgets, providing a broader empirical characterization of federated tabular-to-image learning than prior limited-domain evaluations. 


\end{itemize}

The remainder of this paper is organized as follows. Section~\ref{sec:related_work} reviews related work in FL and tabular-to-image encoding methods. Section~\ref{sec:fed_igtd} describes the Fed-IGTD. Section~\ref{sec:results} presents the datasets used, the experimental setup, and main results. Section~\ref{sec:discussion} discusses limitations and directions for future work, and Section~\ref{sec:conclusions} presents conclusions.

\section{Related work}
\label{sec:related_work}

\subsection{Federated learning}

FL enables a shared global model to be trained across decentralized data holders by exchanging model updates rather than raw data~\cite{zhang2021survey}. In horizontal FL, clients share an identical feature schema but hold different samples, whereas in vertical FL, clients hold partially overlapping features over the same entities. These approaches require different training and aggregation strategies. In horizontal FL, clients independently train models on their local data and contribute model updates to a global aggregation procedure. In contrast, vertical FL typically requires entity alignment (e.g., via private set intersection) and split-model training across the partitioned feature space. Fed-IGTD operates in the horizontal FL setting.

FedAvg is the standard aggregation algorithm for horizontal FL. At each round, a subset of clients initializes local training from the current global model, and the server aggregates the resulting local models, typically using a sample-size-weighted average, to obtain the global model~\cite{mcmahan2017learning}. Although FedAvg was originally analyzed under the IID assumption, practical FL systems often exhibit statistical heterogeneity, as clients' data reflect distinct local populations (e.g., hospital patient populations or regional customer bases). This non-IID setting has motivated methods such as FedProx, which introduces a proximal term that penalizes local model deviation from the global model, improving training stability under heterogeneous data~\cite{li2020federated}. Beyond statistical heterogeneity, prior work has identified architectural limitations of parameter-only FedAvg. In particular, FedBN showed that standard batch normalization is problematic under FedAvg because its mean and variance are non-learnable buffers and are therefore excluded from parameter aggregation, preventing the global model from appropriately updating its normalization statistics~\cite{li2021fedbn}.

FL does not provide formal privacy guarantees. Although raw data remain decentralized, the model updates exchanged during training may still disclose information about individual training samples, making FL vulnerable to attacks such as membership inference, inference of properties associated with training data among others~\cite{chen2024federated}. DP provides formal privacy by perturbing data or model computations with calibrated noise, thereby reducing the risk of revealing information about individual users~\cite{chen2024federated}. DP-SGD applies this principle at the sample level by clipping and perturbing individual gradients during centralized training~\cite{chua2024scalable}. Since DP-SGD releases a noisy statistic at each minibatch step, naive sequential composition over thousands of iterations can result in excessively loose $\varepsilon$ bounds. The moments accountant and its Rényi DP (RDP) formulation enable tighter privacy accounting across repeated releases, with the accumulated privacy loss converted to a standard $(\varepsilon,\delta)$ guarantee at the end of training~\cite{mironov2017renyi}. DP-FedAvg extends this framework to provide client-level privacy in FL. Each client's model update is clipped to a bounded norm, and Gaussian noise is added to the aggregated update before it is applied to the global model, providing privacy at the level of client participation rather than individual records~\cite{mcmahan2017learning}. We adopt the client-level threat model of DP-FedAvg.

\subsection{Tabular-to-image encoding methods}

Tabular-to-image methods aims to construct a feature-to-pixel mapping that preserves the underlying feature relationships by placing related features to spatially close pixels in the resulting image~\cite{liu2025comprehensive}. These approaches employ linear and nonlinear dimensionality reduction techniques, such as Principal Component Analysis (PCA), t-distributed Stochastic Neighbor Embedding (t-SNE), Uniform Manifold Approximation and Projection (UMAP), Multidimensional Scaling (MDS), to embed features into a two-dimensional space from which pixel coordinates are derived. For instance, DeepInsight uses t-SNE, PCA and kernel PCA to project features into a two-dimensional space, which is then rasterized into a fixed-size image grid shared across all samples~\cite{sharma2019deepinsight}. REFINED uses MDS and a hill-climbing algorithm to place correlated features near one another~\cite{bazgir2020representation}. TINTO uses PCA and t-SNE to determine feature locations and the resulting coordinates are then transposed, scaled, and rounded to integer values~\cite{castillo2023tinto}.


Unlike dimensionality reduction methods, IGTD directly optimizes a feature-to-pixel assignment through iterative pairwise swaps, minimizing the discrepancy between the rank orderings of feature-wise and pixel-wise distances on the target grid~\cite{zhu2021converting}. Its rank-based objective depends on the data only through a correlation matrix, which can be reconstructed exactly from additive per-client statistics. This structure yields a relatively simple and computationally efficient objective, making IGTD particularly efficient to federated optimization compared with the non-convex, iterative objectives of methods such as t-SNE and UMAP. IGTD is among the most widely used tabular-to-image encoding methods, having been applied across domains such as cybersecurity~\cite{nazarri2023generating, yzzogh2025evaluating}, the automotive industry~\cite{hosseini2023knock}, and biomedical engineering~\cite{poonam2025hierarchical, lara2025transfer, sonmez2026comparative, wang2026diabetes, gomez2026tabular}. Despite these benefits, a decentralized version of IGTD that provides formal privacy guarantees has not been systematically investigated.

The combination of FL with tabular-to-image conversion is, to our knowledge, addressed by a single prior study~\cite{yuan2026fedvi}. The authors proposes FedVI, a FL module within a broader cross-domain financial risk-modeling framework~\cite{yuan2026fedvi}. It uses vertical FL, in which clients hold different, only partially overlapping feature schemas for the same entities (\textit{e.g.,} a bank and a credit bureau holding different attributes of the same customers). A dual-stream schema-alignment mechanism combines semantic and statistical similarity measures to align heterogeneous feature spaces before constructing a shared image representation.


FedVI reconstructs the correlation matrix used as input to its IGTD layout step through Gram-matrix aggregation, using a mechanism broadly similar to the horizontal approach presented in this study. However, in its vertical, heterogeneous-schema setting, the reconstruction is approximate under non-IID data, with no formal error bounds provided. Additionally, FedVI does not employ DP or provide noise calibration and privacy-loss accounting. Hence, FedVI's privacy analysis is heuristic and does not provide formal privacy guarantees. It relies on the assumption that recovering the original sample matrix from the aggregated Gram matrix is infeasible when the feature dimension is substantially smaller than the number of samples.

\section{Fed-IGTD}
\label{sec:fed_igtd}

Fed-IGTD is a federated framework for transforming tabular data into a shared image representation composed of two stages: \textit{mapping image construction} and \textit{federated training}.

\begin{figure*}
    \centering
    \includegraphics[width=0.99\textwidth]{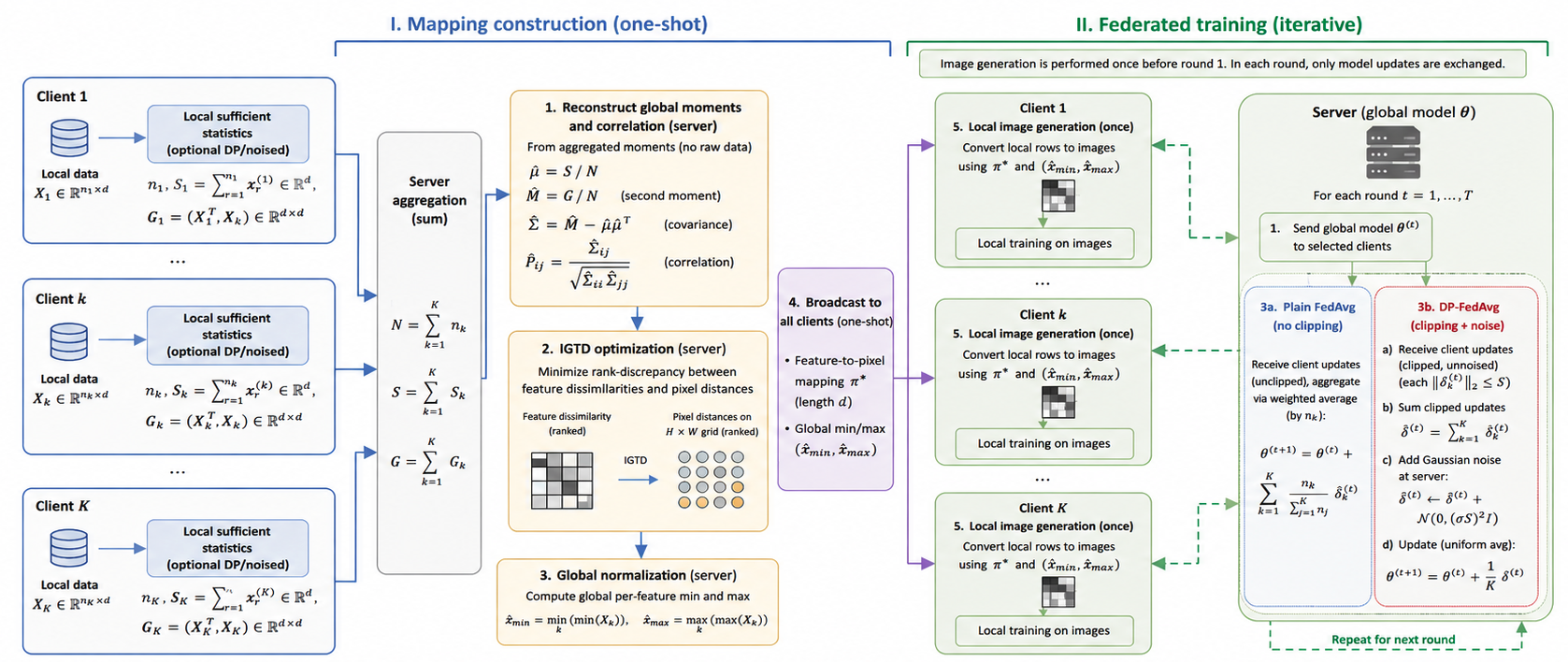} 
    \caption{Fed-IGTD workflow.}
    \label{fig:fed_igtd}
\end{figure*}

In the first stage, clients communicate only aggregate statistics from which the server reconstructs the global feature relationships required by IGTD. The server then computes a single feature-to-pixel mapping and global feature-normalization parameters and broadcasts them to all clients. In the second stage, clients generate images once using the shared mapping and normalization, and train a CNN using either standard FedAvg or DP-FedAvg. In each round, the server sends the global model, receives client updates, and updates the model using either FedAvg (weighted average, no clipping) or DP-FedAvg (clip locally, server adds Gaussian noise to the sum; uniform average). A schematic diagram of Fed-IGTD is shown in Figure~\ref{fig:fed_igtd}.

\subsection{Problem setup}

We consider a federated learning system with $K$ clients. Client $k$ holds a local dataset $X_k \in \mathbb{R}^{n_k \times d}$, consisting of $n_k$ samples and $d$ features. We denote the $r$-th sample of client $k$ by $x_{kr}\in\mathbb{R}^{d}$. The local datasets form a disjoint row-wise partition of the pooled dataset as follows:
$$
X = [X_1^\top,\ldots,X_K^\top]^\top \in \mathbb{R}^{N\times d},
$$
where $N=\sum_{k=1}^K n_k$ and samples $x_{kr}$ are never sent to the server or to other clients. Fed-IGTD aims to construct a shared feature-to-pixel assignment $\phi^*:\mathcal{F}\rightarrow\mathcal{P}$, where 
$$
\mathcal{F}=\{1,\ldots,d\}
$$ 
$$
\mathcal{P}=\{(h,w)\mid h\in\{1,\ldots,H\},\,
w\in\{1,\ldots,W\}\}
$$ 
denote the feature indices and the spatial locations of an $H\times W$ image, respectively. The mapping $\phi^*$ is optimized via an iterative pairwise-swap process by assigning each feature to a unique pixel location while preserving the pairwise relationships between features in the input data. Once the mapping is constructed, each client independently converts its local samples into images and uses them to train a CNN through federated learning. To derive a globally consistent feature-to-pixel mapping without centralizing input data, Fed-IGTD leverages the fact that the feature correlations required by IGTD can be exactly reconstructed from additive statistics computed independently across clients.

\subsection{Reconstruction of feature relationships}

IGTD constructs an image layout by comparing pairwise feature dissimilarities with pairwise spatial distances on the target image grid. Let $P\in\mathbb{R}^{d\times d}$ denote the Pearson correlation matrix, with $P_{ij}=\rho_{ij}$ and define the corresponding feature dissimilarity matrix
$Q \in \mathbb{R}^{d\times d}$, $Q_{ij}=1-|\rho_{ij}|$. Thus, strongly positively or negatively correlated features have small dissimilarity, whereas weakly correlated features have larger dissimilarity.

The key observation underlying Fed-IGTD is that the Pearson correlation matrix can be reconstructed from additive statistics, without sharing input samples. Rather than computing $P$, each client $k$ computes the feature-sum vector $S_k$ and the Gram matrix $G_k$:

$$
S_k=\sum_{r=1}^{n_k}x_{kr}\in\mathbb{R}^d
$$

$$
G_k=X_k^\top X_k\in\mathbb{R}^{d\times d}
$$
where $n_k$ corresponds to its sample count and the server obtains the corresponding pooled statistics by summation:
$$
S=\sum_{k=1}^{K}S_k
$$
$$
G=\sum_{k=1}^{K}G_k
$$
where $N=\sum_{k=1}^{K}n_k$. Using these values, the server reconstructs the global mean $\hat{\mu}$, the second-moment matrix $\hat{M}$, and the reconstructed covariance matrix $\hat{\Sigma}$ as:
$$
\hat{\mu}=\frac{S}{N}
$$
$$
\hat{M}=\frac{G}{N}
$$
$$
\hat{\Sigma}=\hat{M}-\hat{\mu}\hat{\mu}^{\top}
$$
The reconstructed Pearson correlation matrix $\hat{P}$ is then obtained element-wise as follows:
$$
\hat{P}_{ij}
=
\frac{\hat{\Sigma}_{ij}}
{\sqrt{\hat{\Sigma}_{ii}\hat{\Sigma}_{jj}}},
\qquad
i,j=1,\ldots,d,
$$

Finally, we obtain the reconstructed dissimilarity matrix:
$$
\hat{Q}_{ij}=1-\lvert\hat{P}_{ij}\rvert.
$$

Note that $\hat{P}$ coincides with the Pearson correlation matrix that would be computed from centralized data. This means that the reconstruction is independent of the degree of statistical heterogeneity in the client partitions. This property is relevant for Fed-IGTD because an exact reconstruction of $P$ implies that the server can reproduce the same feature-to-pixel layout that would be obtained from centralized data, without requiring any further access to client data. Thus, non-IID label or feature skew across clients has no effect on the correlation-matrix reconstruction, and therefore no effect on the IGTD-derived image layout itself.

\subsection{IGTD layout construction}

Given the reconstructed dissimilarity matrix $\hat{Q}$, the server performs the IGTD optimization without further communication with the clients. Let $D\in\mathbb{R}^{HW\times HW}$ denote the fixed Euclidean distance matrix between all positions on the $H\times W$ image grid. For a candidate feature-to-pixel assignment $\phi$, the spatial distance between features $i$ and $j$ is $D_{\phi(i),\phi(j)}$.

IGTD seeks an assignment for which the ranking of feature dissimilarities agrees with the ranking of their corresponding spatial distances. The optimization is performed following a swapping-based procedure as follows:  
$$
\phi^*
=
\arg\min_{\phi}
\sum_{i<j}
\left|
\operatorname{rank}(\hat{Q}_{ij})
-
\operatorname{rank}
\left(D_{\phi(i),\phi(j)}\right)
\right|.
$$
Starting from an initial assignment $\phi_0$, candidate feature positions are iteratively exchanged. A proposed swap is accepted when it strictly reduces the rank-discrepancy objective; otherwise, the current assignment is retained. The optimization terminates after a predefined iteration budget or after a specified number of consecutive non-improving proposals.

When the number of features $d$ is smaller than the number of grid positions $H{\times}W$ ($d<HW$), the distance matrix is padded with dummy features at maximal distance from all real features, ensuring every grid position receives an assignment while dummy positions are pushed to the periphery of the layout. The final mapping $\phi^*$ is then broadcast to every client as a feature-to-pixel lookup table. The communication cost of this one-shot broadcast is $O(d)$, independent of the number of local samples and of the number of subsequent federated training rounds. An important consequence is that the computationally expensive IGTD optimization is performed only once at the server. Clients do not independently optimize their layouts and therefore cannot produce inconsistent image representations as a consequence of local sample distributions. Every client maps the same feature to the same pixel location.

\subsection{Feature normalization and local image construction}

IGTD image construction requires each feature to be normalized to a common range (typically between 0 and 1) Before feature values are assigned to pixels. In Fed-IGTD, each client reports its local per-feature minimum $m_{k}^{\min}=\min_r x_{kr}$ and maximum $m_{k}^{\max}=\max_r x_{kr}$ (taken independently for each feature). The server reconstructs the global extrema through element-wise aggregation:
$$
\hat{x}_{\min}
=
\min_{k}m_{k}^{\min}
$$
$$
\hat{x}_{\max}
=
\max_{k}m_{k}^{\max}
$$

Each client independently applies the shared assignment $\phi^*$ and global normalization parameters $\hat x_{\min}$ and $\hat x_{\max}$ to convert its local samples into images. For a sample $x\in\mathbb{R}^{d}$, the normalized value of feature $i$ is assigned to the pixel specified by $\phi^*$:
$$
\operatorname{Img}(x)_{\phi^*(i)}
=
z_i(x)
=
\operatorname{clip}
\left(
\frac{x_i-\hat{x}_{\min,i}}
{\hat{x}_{\max,i}-\hat{x}_{\min,i}},
0,1
\right)
$$

Pixels corresponding to dummy features are assigned a fixed background value. Image construction is performed \emph{once}, before the first federated training. The resulting local image datasets are then reused throughout the subsequent training rounds. Consequently, the per-round communication consists only of model parameters or model updates. Note that the image representation itself is not repeatedly transmitted or reconstructed.

\subsection{Federated CNN training}
\label{subsec:fed_training}

The locally generated images are used to train a CNN classifier. At federated round $t$, the server broadcasts the current global model $\theta^{(t)}$ to the selected clients. Each client performs local optimization on its image dataset and obtains a model update:
$$
\delta_k^{(t)}
=
\theta_k^{(t)}-\theta^{(t)}
$$

For FedAvg, the server aggregates client updates using weights proportional to the number of local training samples. For a participating set $\mathcal{K}_t$, the global update is:
$$
\theta^{(t+1)}
=
\theta^{(t)}
+
\sum_{k\in\mathcal{K}_t}
\frac{n_k}
{\sum_{j\in\mathcal{K}_t}n_j}
\delta_k^{(t)}
$$
No clipping or additional noise is required for standard FedAvg. When formal privacy protection of the model-training process is required, Fed-IGTD can use client-level DP-FedAvg (described below). In DP-FedAvg, each client update is first clipped locally to a fixed $L_2$ norm bound $S$:
$$
\tilde{\delta}_k^{(t)}
=
\delta_k^{(t)}
\min
\left(
1,
\frac{S}{\lVert\delta_k^{(t)}\rVert_2}
\right)
$$
The clipped updates are transmitted to the server without noise. The server then sums the clipped updates, adds Gaussian noise to the aggregate, and divides by the number of participating clients:
$$
\theta^{(t+1)}
=
\theta^{(t)}
+
\frac{1}{|\mathcal{K}_t|}
\left[
\sum_{k\in\mathcal{K}_t}
\tilde{\delta}_k^{(t)}
+
\mathcal{N}
\left(
0,(\sigma S)^2I
\right)
\right]
$$

Unlike FedAvg, this DP-FedAvg aggregation uses uniform averaging rather than weights proportional to $n_k$. This ensures that each participating client contributes a bounded amount to the aggregate and provides the fixed sensitivity required for client-level differential privacy. Thus, the two training modes differ only in their model-update aggregation mechanism. Plain FedAvg receives un-clipped client updates and performs sample-size-weighted averaging, whereas DP-FedAvg clips each client update and adds Gaussian noise to the aggregate before applying a uniform average.

\subsection{Privacy approaches}

Fed-IGTD's federated aggregation never transmits raw rows, but statistics such as a Gram matrix are not themselves anonymous: an aggregated Gram matrix, especially with few clients or small $n_k$, can still leak information about the underlying data distribution. Hence, we provide two complementary mechanisms for differential privacy: \textit{(i)} a Gaussian differential privacy mechanism protecting the correlation-matrix aggregation step that determines the IGTD image layout; and \textit{(ii)} a client-level DP-FedAvg mechanism protecting the CNN training process itself across all federated rounds. 
 
\subsubsection{Differential privacy for correlation aggregation}

To bound the sensitivity of the statistics $S_k$ and $G_k$ to any single
training sample, each client clips every sample of $X_k$ to a fixed $L_2$ norm bound
$c$ as follows:
$$
\tilde x_r = x_r \cdot \min\!\left(1,\; \frac{c}{\lVert x_r \rVert_2}\right)
$$
This clipping bounds the $L_2$ sensitivity of the released statistics to the replacement of a single sample: $\Delta S = 2c$ for the sum vector, and
$\Delta G = 2c^2$ for the Gram matrix. Each client then adds independent Gaussian noise, calibrated via the standard Gaussian mechanism, to its local statistics before upload:
$$
\sigma_S = \frac{\Delta S}{\varepsilon/2} \sqrt{2 \ln(1.25/\delta)}
$$
$$
\sigma_G = \frac{\Delta G}{\varepsilon/2} \sqrt{2 \ln(1.25/\delta)}
$$
with the total privacy budget $\varepsilon$ split evenly between the two released statistics via basic sequential composition, and the noised Gram matrix symmetrized. This yields local $(\varepsilon,\delta)$-differential privacy at the row level for the correlation-matrix construction step: an adversary observing the aggregated, noised statistics cannot determine, beyond the guarantee afforded by $\varepsilon$, whether any particular row was present in any client's dataset. 

\subsubsection{Client-level differential privacy FedAvg with R\'enyi-DP accounting}
 
To protect the CNN training process, we adapt DP-FedAvg~\cite{mcmahan2017learning}, which enforces \emph{client-level} rather than row-level privacy: the unit of protection is whether a given institution participated in training, not whether any individual row within that institution's data was used. Each participating client's model update $\delta_k^{(t)}$ is clipped to a fixed $L_2$ norm bound $S$, the server sums the clipped updates, adds Gaussian noise scaled to the clip bound, and applies the noised average to the global model.

Because every client's clipped update has $L_2$ norm at most $S$, the sensitivity of the \emph{sum} $\sum_k \tilde\delta_k^{(t)}$ to any single client joining, leaving, or arbitrarily changing its update is exactly $S$ -- a clean, fixed bound independent of $K$ or of any individual client's local dataset size, which is precisely the property client-level DP requires.

Each round constitutes one release of the Gaussian mechanism with noise multiplier $\sigma$. Rather than composing privacy loss across the $T$ rounds via naive sequential composition (which multiplies the per-round $\varepsilon$ by $T$ and yields prohibitively loose bounds for $T$ in the tens of rounds) we track privacy loss in R\'enyi-divergence space using an R\'enyi-differential-privacy (RDP) accountant, composing the $T$ per-round Gaussian-mechanism RDP curves
additively and converting to a single $(\varepsilon,\delta)$ guarantee only after all $T$ rounds have completed:
$$
\varepsilon = \operatorname{RDP\text{-}to\text{-}DP}\!\left(
\sum_{t=1}^{T} \operatorname{RDP}_{\mathcal{N}(0,\sigma^2)}(\alpha),\; \delta\right)
$$
minimized over the R\'enyi order $\alpha$. Because our simulated federation uses a fixed cohort of $K=5$ clients, all of whom participate in every round (no random per-round subsampling from a larger population), we set the accountant's sample rate to $1.0$, honestly reflecting the absence of any privacy amplification via subsampling; in an extension to a larger cross-silo population with random per-round participation, this amplification could be claimed to further tighten $\varepsilon$ for the same
$\sigma$.

\section{Results}
\label{sec:results}

In this section, we present the datasets used, experimental setup as well as the classification results using Fed-IGTD.

\subsection{Datasets}
\label{sec:dataset_description}

The evaluated datasets are sourced from public repositories, spanning five domains: healthcare (Pima Indians Diabetes), finance (German Credit), chemistry/agriculture (Wine Quality), signal processing (Sonar), and physics/radar (Ionosphere). Table~\ref{table:dataset_summary} summarizes the datasets, including the number of samples and features.

\begin{table}[htbp]
\centering
\caption{Datasets used in this study.}
\label{table:dataset_summary}
\scalebox{0.95}{
\begin{tabular}{lrr}
\toprule
Dataset                          & Samples  & Features  \\ \midrule
Pima Indians Diabetes            & 768  & 8                 \\
German Credit                    & 1000 & 20                \\
Wine Quality (White)             & 4898 & 11                \\
Connectionist Bench (Sonar)      & 208  & 60                \\
Ionosphere                       & 351  & 34 \\
\bottomrule
\end{tabular}
}
\end{table}

For Wine Quality, the original quality score is binarized into two classes: quality $\geq 6$ (good) and quality $< 6$ (not good). For German Credit, categorical features are encoded using ordinal label encoding. For Ionosphere, one constant feature (zero-variance) was removed, consequently, 33 of the original 34 features were used.

\subsection{Experimental setup}
\label{sec:experimental_setup}

\textit{Data splitting.} For each dataset, we hold out 20\% of the samples as a stratified test set (evaluated centrally), with the remaining 80\% partitioned across clients for federated training. We simulate a cross-silo federation of $K=5$ clients, all of whom participate in every communication round. To systematically control non-IID severity, we adopt the commonly used Dirichlet-based label-skew protocol, in which each client's class proportions are sampled from a Dirichlet distribution with concentration parameter $\alpha$. Larger $\alpha$ yields a distribution closer to IID, whereas smaller $\alpha$ produces stronger label skew. This provides a single interpretable parameter for controlling the degree of heterogeneity. We  considered a concentration parameter $\alpha \in \{1000, 1.0, 0.1\}$, corresponding to approximately IID, moderately non-IID, and severely non-IID partitions.

\textit{Federated simulation.} Each client independently computes its local statistics $(n_k, S_k, G_k)$, which are aggregated by the server to reconstruct the global correlation matrix and normalization bounds. The resulting IGTD mapping is optimized via swap-based iterative manner with a maximum of 3000 iterations and a patience threshold of 400 consecutive non-improving proposals. The image grid dimensions $H \times W$ are chosen as the smallest near-square grid satisfying $H \times W \geq d$ for each dataset's feature count $d$.

\textit{Model architecture and training.} Each client trains a compact CNN classifier consisting of two convolutional layers (16 and 32 channels, $3\times3$ kernels) with Group Normalization followed by a fully connected layer (64 units) with dropout ($p=0.3$) and a final classification layer. Models are trained via Adam (learning rate $10^{-3}$) for 15 federated rounds with 1 local epoch per round, using either standard FedAvg (weighted by local sample count, no clipping) or client-level DP-FedAvg. As a centralized reference, we train the same architecture on the pooled training data using the centrally computed IGTD mapping.

\textit{Privacy configuration.} For the one-shot Gram-matrix DP mechanism, we sweep the privacy budget $\varepsilon \in \{8, 2, 0.5\}$ at fixed $\delta = 10^{-5}$, alongside a non-private baseline, holding non-IID severity fixed at $\alpha=1.0$ to isolate the effect of privacy noise from that of client heterogeneity. For DP-FedAvg, we sweep the noise multiplier $\sigma \in \{10^{-4}, 0.05, 0.1, 0.2, 0.4, 0.8, 1.5, 3, 6, 12\}$, reporting the actually-achieved $\varepsilon$ at $\delta=10^{-5}$ as computed by the R\'enyi-DP accountant after composing privacy loss across all 15 training rounds, rather than a nominal, unaccounted noise level. The DP-FedAvg sweep uses the noise-free federated correlation reconstruction to construct the IGTD mapping. DP-FedAvg protects only the federated training, not the mapping-construction step. The R\'enyi-DP accountant evaluates R'enyi orders $\alpha \in {1.1, 1.2, \ldots, 10.9} \cup {11, 12, \ldots, 63}$, comprising 143 orders in total, and reports the tightest resulting $(\varepsilon,\delta)$ bound. The per-client update clip bound $S$ is calibrated per dataset via a short, non-private probe (3 rounds without noise), taking the median observed update norm as a practical estimate.

\textit{Communication cost.} Mapping construction requires each client to upload one Gram matrix ($O(d^2)$) and one sum vector ($O(d)$) in a single round, after which the server broadcasts the resulting mapping as an $O(d)$ lookup table plus $O(d)$ normalization bounds -- a fixed, one-time cost independent of subsequent training. Federated training then transmits one full model parameter vector per client per round, for 15 rounds, regardless of $d$; for the model architecture used here (Section~\ref{sec:experimental_setup}), this per-round cost is several orders of magnitude larger than the one-time $O(d^2)$ mapping-construction cost for every dataset in our suite, since $d \leq 60 \ll$ the number of model parameters.

\textit{Evaluation.} We report two metrics: test accuracy and macro-averaged F1 score, the latter weighting all classes equally and therefore more sensitive to degradation on underrepresented classes. All federated and centralized results are averaged over 5 independent seeds (3 for Wine Quality, due to its larger sample count and correspondingly higher per-run training cost), each seed governing the train/test split, the Dirichlet client partition, and model initialization jointly. We report mean and standard deviation across seeds.

\subsection{IGTD correlation matrix using centralized and federated approach}

\begin{figure*}
  \centering
  \begin{subfigure}[b]{0.85\linewidth}
    \includegraphics[width=\linewidth]{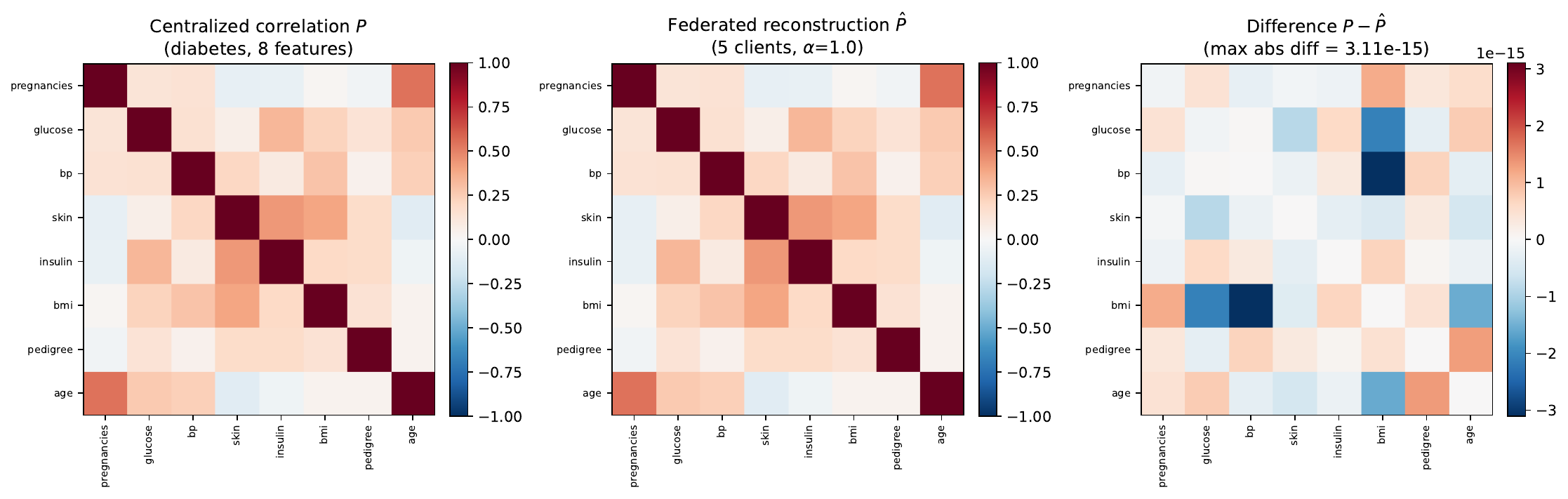} 
    \caption{} 
  \end{subfigure}
  \begin{subfigure}[b]{0.8\linewidth}
    \includegraphics[width=\linewidth]{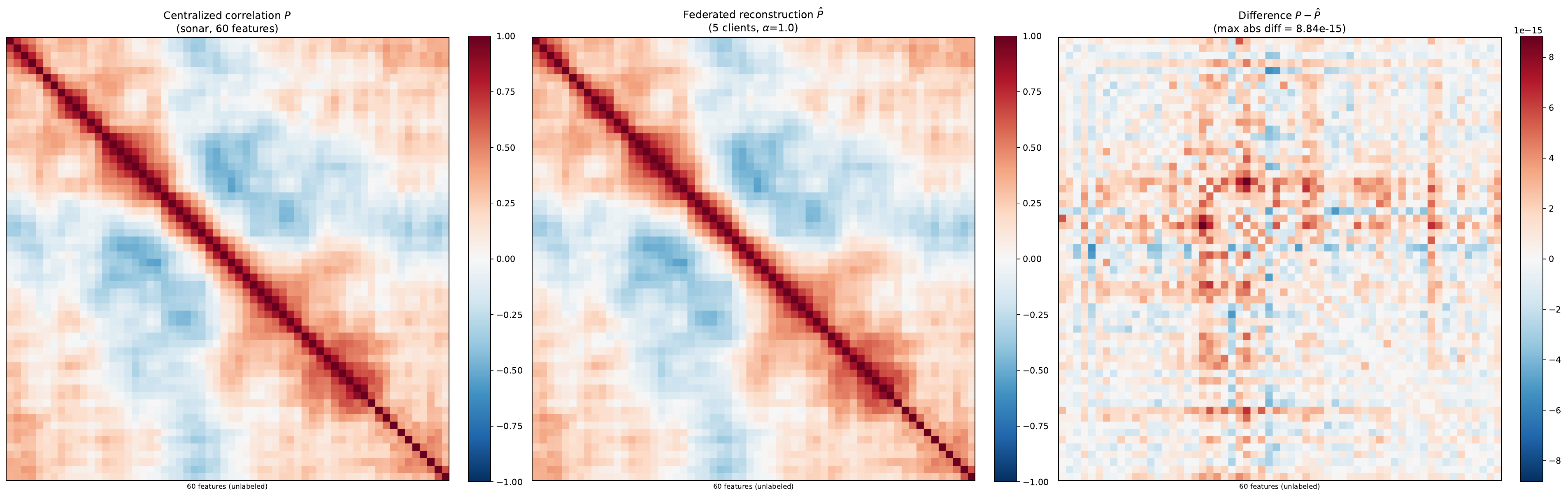} 
    \caption{} 
  \end{subfigure}
  \captionsetup{justification=justified, singlelinecheck=false, margin=6pt} 
  \caption{Comparison of centralized and federated correlation matrix. Considering the dataset of: (a) Diabetes (8 features); and (b) Sonar (60 features).}
  \label{fig:heatmap_comparison} 
\end{figure*}

Figure~\ref{fig:heatmap_comparison} compares the centrally computed IGTD correlation matrix $P$ with its federated reconstruction $\hat P$ for two datasets corresponding to the minimum and maximum feature dimensionalities in our evaluation: Diabetes ($d=8$) and Sonar ($d=60$). For both datasets, $P$ and $\hat P$ are visually indistinguishable, while the difference matrix $P-\hat P$ exhibits no discernible structure, with residuals on the order of $10^{-15}$. These differences are several orders of magnitude smaller than any correlation discrepancy expected to affect the resulting IGTD layout and are consistent with floating-point round-off rather than systematic reconstruction error. The correlation matrices for the rest of datasets are shown in Figure~\ref{fig:heatmap_comparison_extension} of Appendix A. The same behavior was observed across all datasets, with maximum absolute differences ranging from $1.98\times10^{-15}$ for German Credit to $1.68\times10^{-11}$ for Wine Quality. The slightly larger residual for Wine Quality is attributable to accumulated floating-point error associated with its substantially larger sample size, rather than to approximation in the federated reconstruction. These results confirm that $\hat P$ reproduces $P$ to numerical precision across the evaluated datasets, independently of sample size and feature dimensionality.

\subsection{Centralized versus federated predictive performance}

\begin{figure}[!hbtp]
  \centering
  \begin{subfigure}[b]{0.95\linewidth}
    \includegraphics[width=\linewidth]{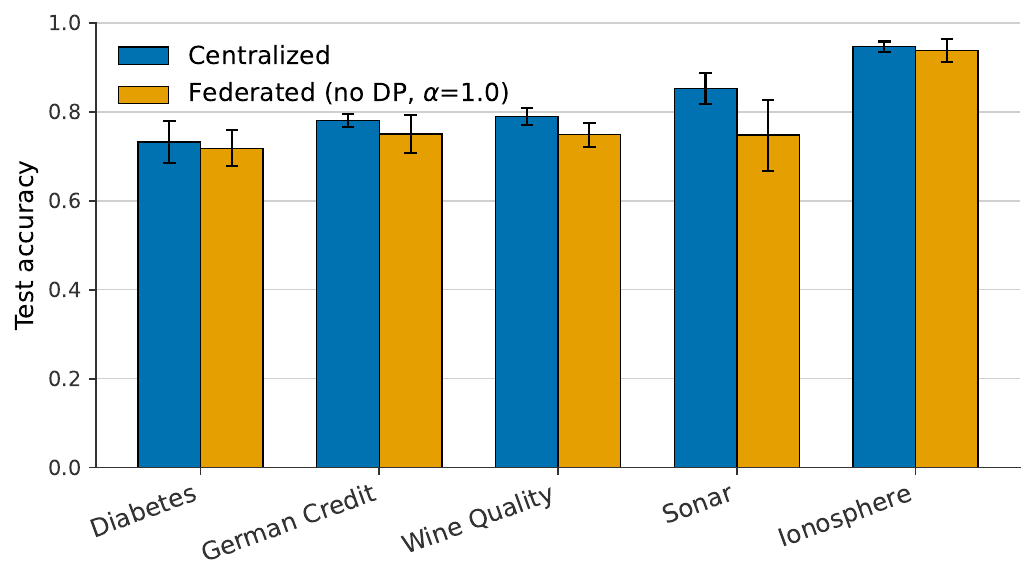} 
    \caption{} 
  \end{subfigure}
  \begin{subfigure}[b]{0.95\linewidth}
    \includegraphics[width=\linewidth]{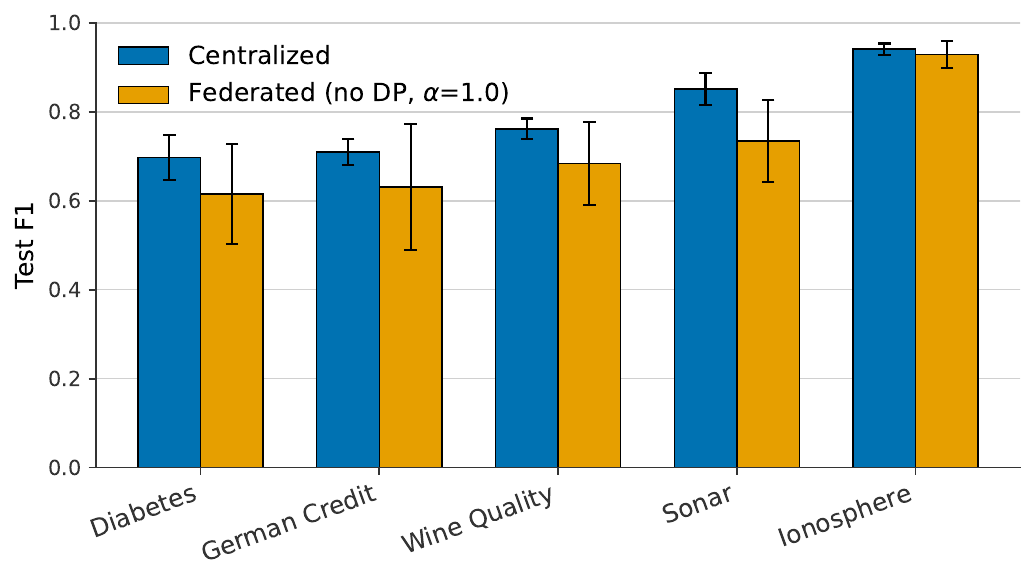} 
    \caption{} 
  \end{subfigure}
  \captionsetup{justification=justified, singlelinecheck=false, margin=6pt} 
  \caption{Centralized and federated predictive performance under FedAvg with five clients ($\alpha=1.0$, without differential privacy). (a) Test accuracy; and (b) Test macro-F1.}
  \label{fig:centralized_vs_federated_performance} 
\end{figure}

Figure~\ref{fig:centralized_vs_federated_performance} compares centralized and federated predictive performance under FedAvg across five clients with moderate label skew ($\alpha=1.0$) and without DP, establishing a baseline prior to the introduction of more severe non-IID heterogeneity or privacy perturbation. As shown in Figure~\ref{fig:centralized_vs_federated_performance}~(a), federated training yields consistently lower accuracy than the centralized reference across most datasets, with decreases ranging from approximately two to ten percentage points for Diabetes (from 0.734 to 0.716), German Credit (from 0.780 to 0.749), Wine Quality (from 0.789 to 0.748), and Sonar (from 0.851 to 0.748). In contrast, Ionosphere demonstrates greater robustness to federated training, with accuracy decreasing only from 0.947 to 0.938. It is also the only dataset for which the confidence intervals of the centralized and federated estimates exhibit substantial overlap, indicating limited performance loss under federation.

The macro-F1 values in Figure~\ref{fig:centralized_vs_federated_performance}~(b) reveal a larger centralized--federated performance gap than is apparent from accuracy alone, particularly for Sonar and German Credit. This finding reflects the greater sensitivity of macro-F1 to class-specific performance degradation. Unlike accuracy, macro-F1 assigns equal weight to each class, preventing errors concentrated in minority classes from being masked by the performance of the majority class. Ionosphere again exhibits the smallest and most stable centralized--federated discrepancy across both metrics, suggesting that the impact of federated training is inherently dataset-dependent.

\subsection{Effect of non-IID severity across domains}

We first characterize the three federation severity levels (IID, Moderate, and Severe) at the client level. Figure~\ref{fig:dirichlet_supp} in Appendix B shows the resulting Dirichlet label-skew partitions across all datasets, with each row reporting client-level sample counts and class distributions for the three severity levels, together with the mean normalized per-client entropy. Note that 1.0 indicates a perfectly balanced class distribution and 0.0 indicates a single-class distribution. For every dataset, increasing federation severity is accompanied by a monotonic reduction in mean entropy, although the underlying partition characteristics vary across datasets. For instance, German Credit exhibits the most extreme severe partition in terms of mean entropy (0.18), with four of the five clients receiving fewer than 10 samples, effectively concentrating the available training data in a single dominant client. Ionosphere provides a contrasting case, with a relatively high mean entropy of 0.62 despite approximately 85\% of the samples being assigned to a single client, because the dominant client's local class distribution remains comparatively balanced. This highlights an important limitation of mean client entropy as a measure of federation severity, as it captures within-client class balance but does not account for heterogeneity in client-level sample sizes. Consequently, substantial imbalance in sample allocation across clients may coexist with relatively high mean entropy.

\begin{figure*}[!hbtp]
  \centering
  \begin{subfigure}[b]{0.95\linewidth}
    \includegraphics[width=\linewidth]{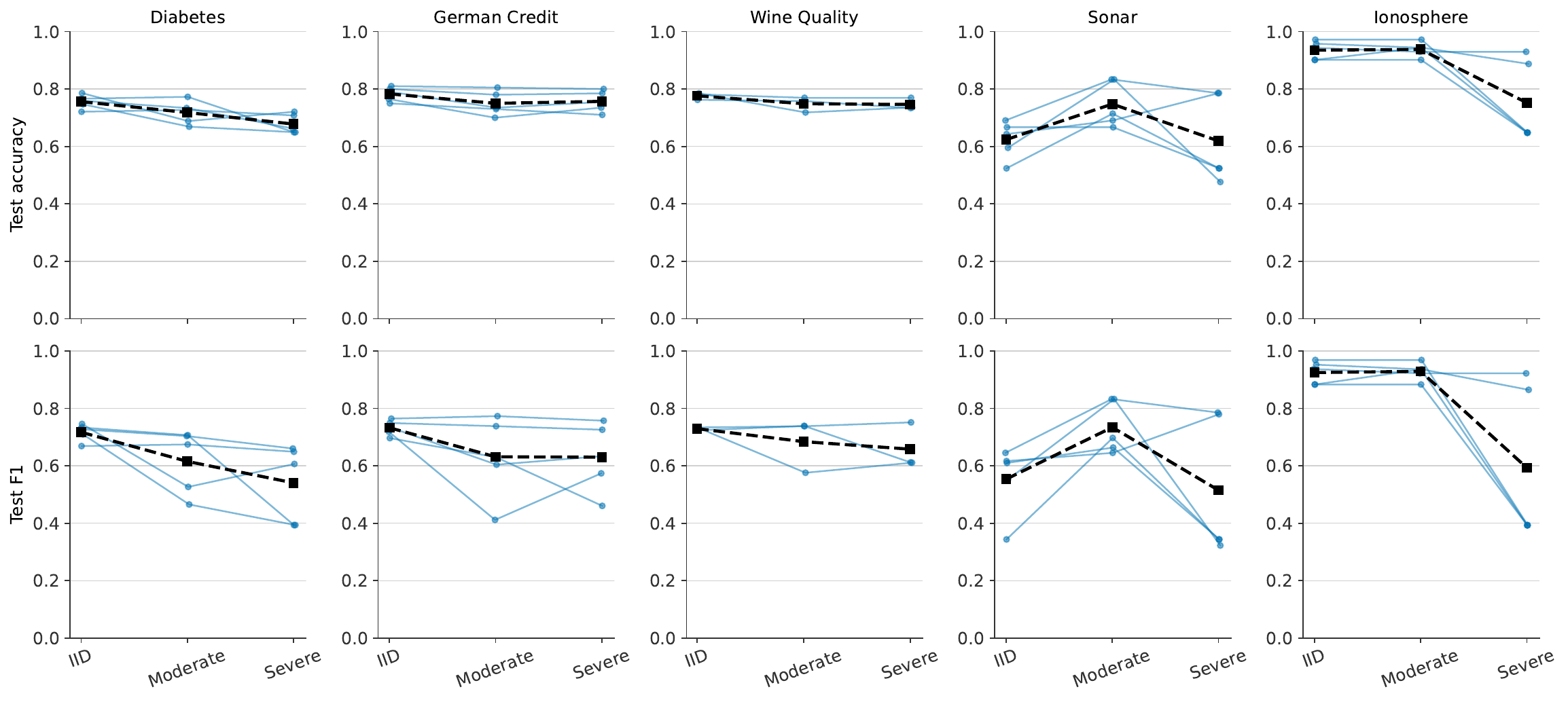} 
  \end{subfigure}
  \captionsetup{justification=justified, singlelinecheck=false, margin=6pt} 
  \caption{Per-seed accuracy (top row) and macro-F1 (bottom row) trajectories across the non-IID severity sweep (IID, Moderate, Severe). Each thin line traces one seed's value across the three settings; the dashed black line marks the mean.}
  \label{fig:non_iid_distributions} 
\end{figure*}

Using these partitions, Figure~\ref{fig:non_iid_distributions} presents the per-seed trajectories of accuracy and macro-F1 across the three federation severity levels. Diabetes, German Credit, and Wine Quality exhibit the expected trend of progressively declining performance as heterogeneity increases, with accuracy trajectories remaining relatively close to their respective means, whereas macro-F1 exhibits greater seed-to-seed dispersion. This variability is particularly pronounced for Diabetes and German Credit under severe skew, where macro-F1 spans approximately 0.4--0.65 despite comparatively tightly clustered accuracy values.

Sonar and Ionosphere exhibit different results to increasing non-IID severity. For Sonar, both accuracy and macro-F1 initially improve from IID to Moderate skew before declining sharply under Severe skew, with individual seed trajectories diverging and crossing between the Moderate and Severe conditions rather than following a common trend. This behavior may be attributable to the limited sample size of Sonar, with only 208 observations distributed across five clients, thereby increasing the sensitivity of the resulting models to random client assignments as per-client sample sizes become increasingly constrained under severe skew. Ionosphere exhibits the most pronounced degradation. Accuracy and macro-F1 remain relatively stable and close to the performance ceiling under IID and Moderate conditions, but diverge substantially under Severe skew, with macro-F1 spanning approximately 0.4--0.9 across seeds. As the severe partition concentrates most samples on a single client while maintaining a comparatively balanced class distribution within that client (see Figure~\ref{fig:dirichlet_supp}), the observed instability appears to be driven primarily by substantial across-client sample-size imbalance rather than within-client class imbalance. This behavior constitutes a distinct failure mode relative to the progressive degradation observed across the other datasets and highlights the limitations of entropy as a standalone measure of federation severity. Under Severe skew, both Sonar and Ionosphere exhibit markedly greater seed-to-seed variability in macro-F1 than in accuracy, consistent with its greater sensitivity to class-specific performance degradation.

\subsection{Effect of differential privacy on the correlation-aggregation step}

\begin{figure*}[!hbtp]
  \centering
  \begin{subfigure}[b]{0.95\linewidth}
    \includegraphics[width=\linewidth]{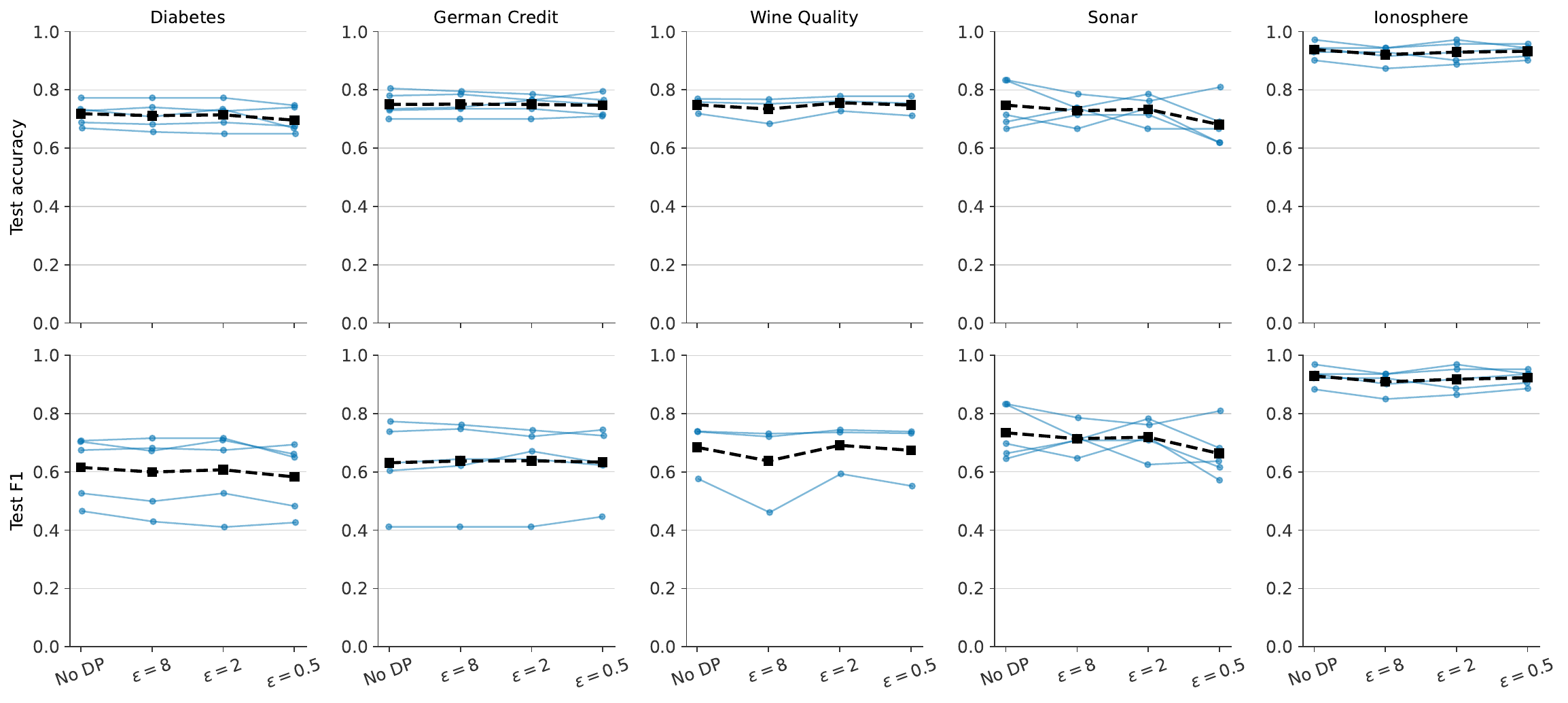} 
  \end{subfigure}
  \captionsetup{justification=justified, singlelinecheck=false, margin=6pt} 
  \caption{Per-seed accuracy (top row) and macro-F1 (bottom row)
    trajectories across the Gram-matrix differential privacy sweep, at
    fixed moderate non-IID severity ($\alpha=1.0$). Each thin line traces one seed's value     across the four privacy budgets (No DP, $\varepsilon=8$, $\varepsilon=2$, $\varepsilon=0.5$); the dashed black line marks the
    mean.}
  \label{fig:dp_distributions} 
\end{figure*}

Figure~\ref{fig:dp_distributions} isolates the utility cost associated with protecting the one-shot correlation-aggregation step used to construct the IGTD image mapping, while fixing the non-IID severity at $\alpha=1.0$. By fixing the degree of non-IID heterogeneity, this configuration isolates the contribution of privacy perturbation to predictive results. In contrast to the degradation observed under severe non-IID conditions, mean accuracy remains stable across all five datasets over the evaluated privacy range, including at $\varepsilon=0.5$. The individual seed trajectories likewise remain closely clustered as the privacy budget decreases, with no substantial increase in between-seed variability. Because the Gram-matrix aggregation step is reconstructed exactly regardless of the noise added to the released statistics, DP noise here perturbs only the IGTD layout itself, a considerably smaller and more localized disturbance than the client-composition changes that drive the instability seen under severe non-IID skew.

Macro-F1 exhibits a similar pattern, with only minor deviations for German Credit and Wine Quality, where a small decline at $\varepsilon=8$ is followed by recovery at tighter privacy budgets ($\varepsilon=2$ and $\varepsilon=0.5$). This non-monotonic variation does not appear to reflect a systematic privacy-induced degradation, as the initial decline does not persist or increase with further reductions in $\varepsilon$. Moreover, none of the datasets exhibits the pronounced seed-to-seed divergence observed for Ionosphere and Sonar under severe non-IID conditions. Notably, even these two datasets, which showed the greatest sensitivity to federation heterogeneity in previous subsection, maintain relatively stable and tightly clustered trajectories under the present privacy setting.

These results indicate that the privacy cost of protecting only the mapping-construction step is small relative to the cost of severe non-IID skew, or client-level DP-FedAvg applied throughout the training process. This difference can be attributed to the distinct points at which privacy noise is introduced. Gram-matrix DP perturbs a single, low-dimensional statistic computed once prior to training, whereas DP-FedAvg perturbs model updates at every training round, allowing the effects of noise to accumulate throughout optimization.

\subsection{Characterizing the privacy–utility tradeoff in client-level DP-FedAvg}

\begin{figure*}[!hbtp]
  \centering
  \begin{subfigure}[b]{0.85\linewidth}
    \includegraphics[width=\linewidth]{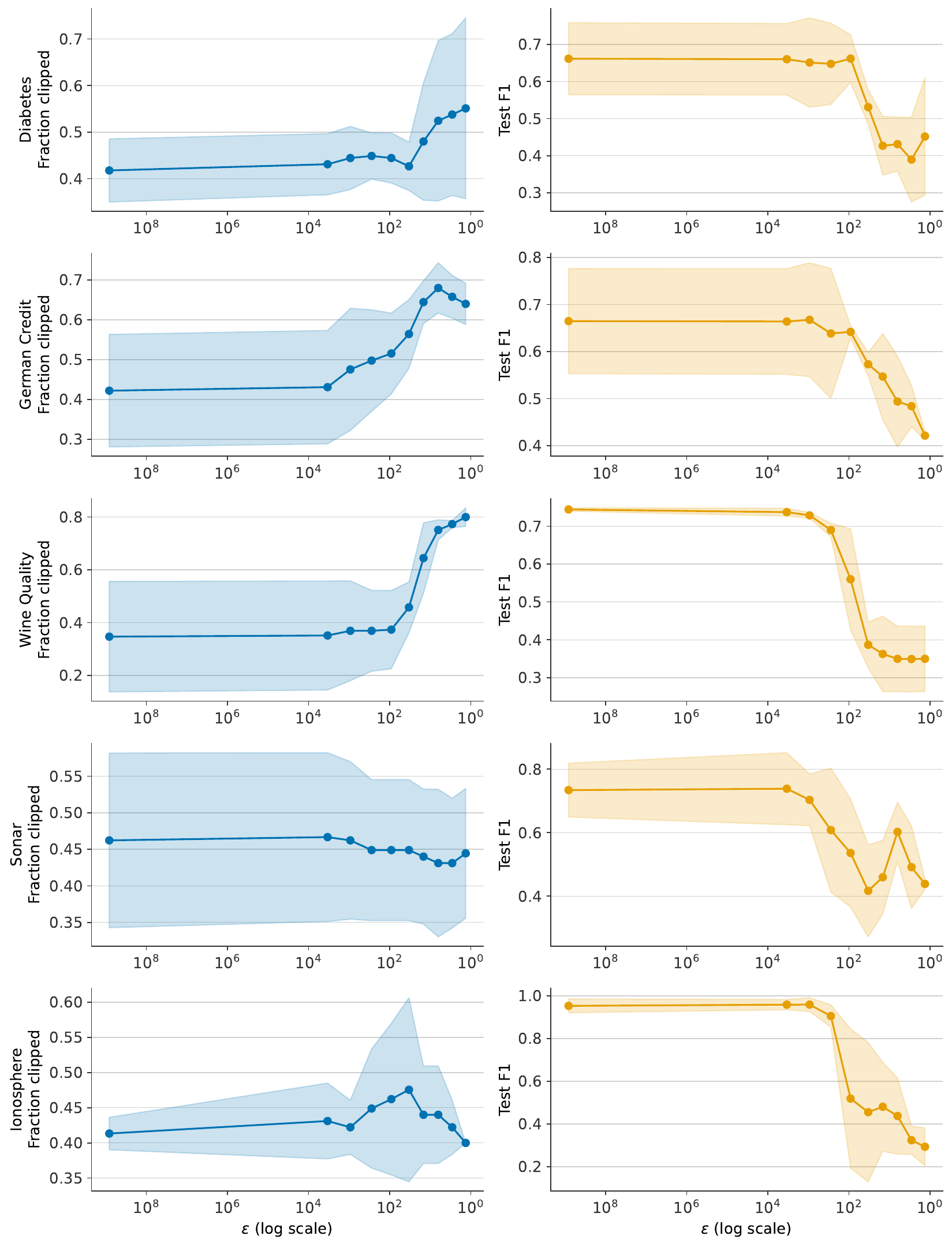} 
  \end{subfigure}
  \captionsetup{justification=justified, singlelinecheck=false, margin=6pt} 
  \caption{DP-FedAvg mechanistic diagnostic across five datasets. Left
    column: fraction of client updates hit by clipping vs.\ achieved
    privacy budget $\varepsilon$ (log scale, decreasing left to right).
    Right column: test macro-F1 vs.\ the same $\varepsilon$, for reference.}
  \label{fig:frac_clipped} 
\end{figure*}

Figure~\ref{fig:frac_clipped} characterizes the DP-FedAvg privacy--utility tradeoff by examining two complementary quantities: the fraction of client updates clipped per round (left) and the resulting test macro-F1 (right), both as a function of the achieved privacy budget $\varepsilon$. For three of the five datasets (Diabetes, German Credit, and Wine Quality), the two quantities exhibit a clear inverse relationship: decreasing $\varepsilon$ is associated with a progressively higher clipping fraction and lower macro-F1. This behavior aligns with the mechanism described above, as noise grows relative to the clip bound $S$, a larger share of client updates are truncated before aggregation, degrading the federated model's effective training signal.

Sonar and Ionosphere, however, deviate from this pattern. For Sonar, the clipping fraction remains nearly constant across the $\varepsilon$ range, decreasing slightly from 0.46 to 0.44 at the tightest privacy budgets, while macro-F1 nevertheless declines sharply from approximately 0.74 to 0.40. Ionosphere exhibits a similar disconnect, with the clipping fraction varying only modestly and non-monotonically, reaching a peak of approximately 0.48 before declining to 0.40, whereas macro-F1 drops from approximately 0.95 to 0.27. For these two datasets, the clipping fraction alone does not fully explain the observed degradation in macro-F1, suggesting that the magnitude of the injected Gaussian noise relative to the model and dataset sizes constitutes an additional determinant of DP-FedAvg utility. This effect appears particularly pronounced for smaller datasets, such as Sonar with 208 samples, as well as for datasets that exhibited comparatively greater instability under severe skew in the non-IID experiments. These results indicate that the clipping fraction is an important, but not sufficient, determinant of the DP-FedAvg utility decline. Clipping appears to be the primary contributor to performance degradation in larger and more homogeneous datasets, while the magnitude of the injected noise relative to model and dataset scale introduces an additional, dataset-dependent effect that is not captured by clipping frequency alone.

\subsection{Privacy–utility tradeoffs across privacy mechanisms}

\begin{figure}[!hbtp]
  \centering
  \begin{subfigure}[b]{0.95\linewidth}
    \includegraphics[width=\linewidth]{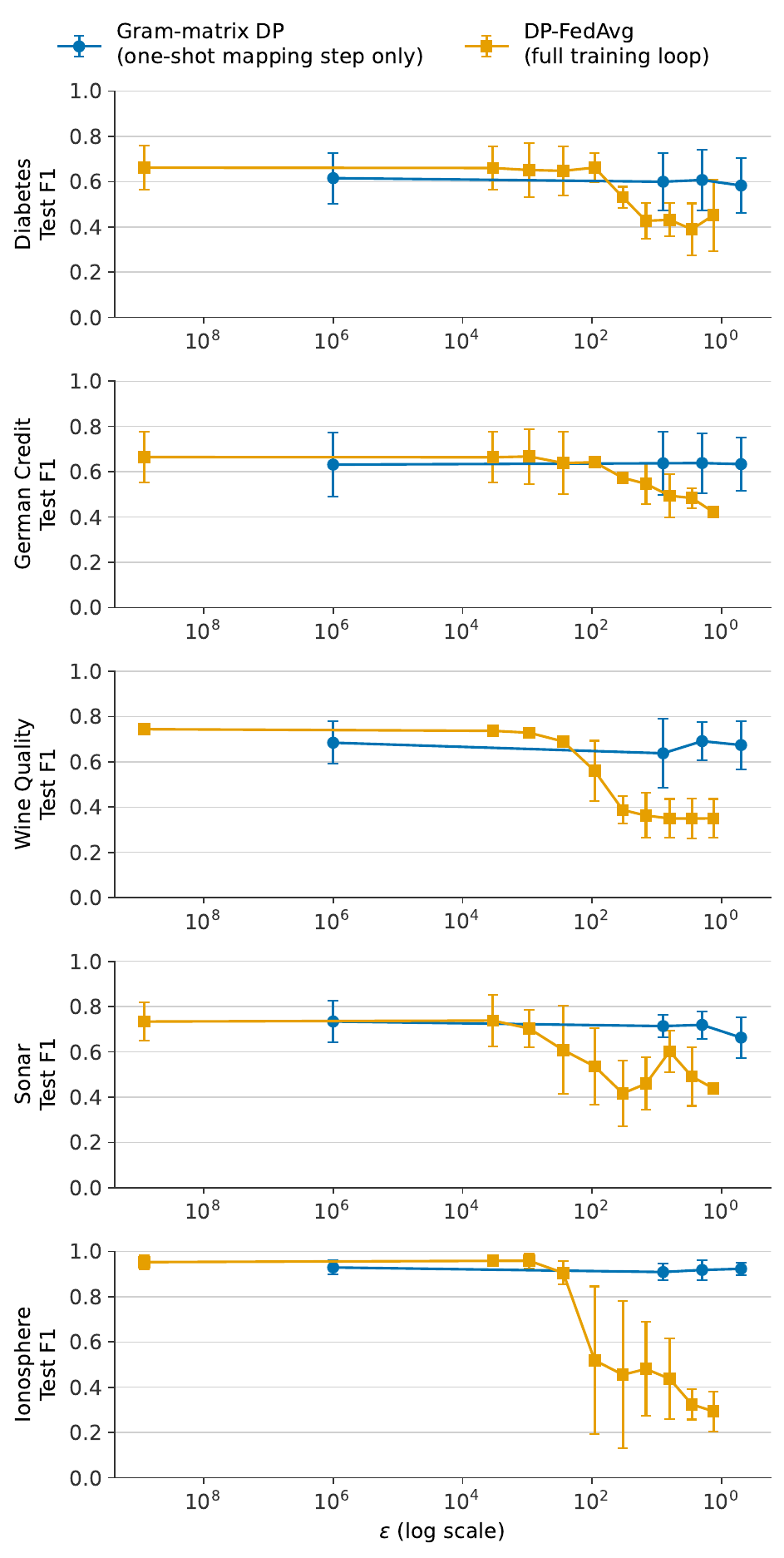} 
  \end{subfigure}
  \captionsetup{justification=justified, singlelinecheck=false, margin=6pt} 
  \caption{Comparison of the two privacy mechanisms on test macro-F1:
Gram-matrix DP protecting only the one-shot mapping-construction step
(blue) versus client-level DP-FedAvg protecting the full training loop
(orange), plotted against the achieved privacy budget $\varepsilon$ (log scale, decreasing left to right) across five datasets.}
  \label{fig:dp_mechanism_comparison} 
\end{figure}

Figure~\ref{fig:dp_mechanism_comparison} provides a direct comparison of the two privacy mechanisms evaluated in this study across all five datasets. Gram-matrix DP (blue) exhibits negligible variation across the range of evaluated privacy budgets, indicating that privacy protection confined to the one-shot correlation-aggregation step incurs minimal utility degradation, even under stringent privacy constraints. In contrast, DP-FedAvg (orange), which applies privacy protection at each round of the federated training procedure, exhibits a pronounced utility decline as $\varepsilon$ decreases across all five datasets. The systematic divergence between the two curves demonstrates that extending privacy protection from a single aggregation step to the entire iterative training process substantially increases the associated utility cost, despite the two mechanisms being evaluated under comparable nominal privacy budgets.

The magnitude of the DP-FedAvg utility degradation is strongly dataset-dependent. Ionosphere exhibits the most pronounced and seed-sensitive decline, with macro-F1 decreasing from near-ceiling performance to a range of approximately 0.3--0.5 under the most stringent privacy budgets. This observation is consistent with the earlier finding that Ionosphere's DP-FedAvg instability is more closely associated with the magnitude of the injected noise than with clipping frequency. In particular, a model operating near ceiling performance may have limited tolerance to perturbations before noise-induced changes substantially affect its decision boundary. Wine Quality and Sonar exhibit similarly pronounced degradation, with macro-F1 declining to approximately 0.35--0.40 at the tightest privacy budgets. In contrast, Diabetes and German Credit exhibit a more moderate decline followed by partial recovery, with the DP-FedAvg confidence intervals overlapping those of Gram-matrix DP across substantial portions of the intermediate-$\varepsilon$ range. These results suggest that, for these datasets, although the direction of the privacy-induced utility loss is consistent, its magnitude remains subject to substantial seed-level variability, particularly given the limited number of random seeds (3) considered in the experiments.

This comparison has direct practical implications for the deployment of Fed-IGTD. A fixed nominal privacy budget $\varepsilon$ does not entail a uniform utility cost, as the resulting performance degradation depends critically on the stage of the pipeline to which privacy protection is applied. Privacy protection of the mapping-construction step alone incurs negligible utility loss and may therefore be considered a default configuration. In contrast, extending privacy protection to the full training procedure through DP-FedAvg introduces a substantially greater utility cost. For several datasets in our evaluation, particularly those that are smaller or more sensitive to perturbations, maintaining practical utility under stringent privacy constraints may consequently require a larger client cohort or a less restrictive privacy target.

\section{Discussion}
\label{sec:discussion}

This paper introduces Fed-IGTD, a horizontal FL-based framework for privacy-preserving tabular-to-image transformation based on IGTD. The proposed approach enables the correlation structure required by IGTD to be reconstructed through federated aggregation, without requiring the underlying tabular data to leave individual clients. Across the five datasets, the results demonstrate substantial dataset dependence in the robustness of Fed-IGTD to federation, statistical heterogeneity, and DP. For instance, Ionosphere, despite achieving near-ceiling performance under centralized and IID settings, exhibits pronounced sensitivity to severe non-IID skew and DP-FedAvg noise, whereas Diabetes and German Credit show comparatively gradual performance degradation. This pattern suggests that models operating near the performance ceiling may have limited tolerance to perturbations, such that relatively small changes in the learned decision boundary can result in substantial losses in predictive performance. Sonar exhibits a different form of sensitivity, which is consistent with its limited sample size. With only 208 samples distributed across five clients, variations in client-level data assignments can have a comparatively large influence on the resulting local data distributions and, consequently, on the global model. The consistency of these patterns across the non-IID severity analysis, the DP-FedAvg clipping analysis, and the direct comparison of privacy mechanisms further indicates that dataset characteristics play a substantive role in determining federated robustness. These findings underscore the importance of reporting dataset scale, class distribution, and baseline performance when evaluating federated tabular-to-image methods, as aggregate predictive performance alone may not adequately characterize robustness across different data regimes.

The privacy experiments also reveal an important distinction between protecting a single preprocessing step and protecting the complete learning procedure. Gram-matrix DP introduces limited utility degradation across the evaluated privacy budgets because privacy protection is applied only to the one-shot correlation-aggregation step preceding model training. In contrast, client-level DP-FedAvg applies perturbation throughout the iterative training process, resulting in substantially greater and more dataset-dependent utility degradation, particularly for the small client cohort ($K=5$) considered in this study. These results motivate a layered approach to privacy protection in practical deployments of Fed-IGTD. When protection of the correlation-aggregation step is sufficient for the intended threat model, Gram-matrix DP offers a comparatively low-cost privacy mechanism. When protection of the complete training process is required, DP-FedAvg provides stronger end-to-end protection but may require additional resources, such as a larger client cohort, or a less stringent privacy target to preserve useful predictive performance. The appropriate mechanism therefore depends not only on the desired privacy budget but also on which components of the pipeline must be protected.

The evaluation also highlights the importance of metric selection when assessing federated and privacy-preserving learning. Across the experiments, macro-F1 was generally more sensitive to degradation than accuracy, particularly under severe non-IID skew and strong DP perturbation. This difference reflects the fact that accuracy can remain relatively stable when performance deteriorates disproportionately on minority classes, whereas macro-F1 gives equal weight to each class. Consequently, accuracy alone may obscure practically important degradation in class-imbalanced federated settings. This consideration is particularly relevant for cross-institutional deployments, where heterogeneous class distributions and limited per-client sample sizes are common. We therefore recommend reporting macro-averaged or per-class metrics alongside accuracy when evaluating federated tabular-to-image methods, especially under distribution shift or privacy constraints.

To our knowledge, FedVI remains the only previously published system that combines FL with tabular-to-image encoding, providing an important point of comparison for Fed-IGTD. The two approaches differ in several fundamental points. First, FedVI adopts vertical FL in which heterogeneous client feature spaces require a dual-stream schema-alignment procedure, resulting in an approximate reconstruction of the correlation matrix under heterogeneous data distributions. In contrast, Fed-IGTD adopts horizontal FL that permits the required correlation structure to be reconstructed, independent of client heterogeneity. 

Second, the privacy treatment differs substantially between the two frameworks. FedVI does not provide a formal DP mechanism or explicit privacy-loss accounting. Fed-IGTD, by contrast, incorporates two formally specified privacy mechanisms with explicit guarantees: Gaussian perturbation of the one-shot correlation-aggregation step and client-level DP-FedAvg for protection throughout the training process. The latter is composed using a R'enyi differential privacy accountant to obtain an explicit $(\varepsilon,\delta)$ guarantee rather than relying on an uncalibrated noise level. Third, whereas the empirical evaluation of FedVI is limited to two financial credit-risk datasets and three participating clients, our evaluation spans five domains, including healthcare, finance, chemistry/agriculture, signal processing, and radar/ physics, and systematically examines both non-IID severity and privacy budgets. This evaluation reveals dataset-dependent robustness patterns that are not apparent from a single-domain setting. 

\section{Conclusions}
\label{sec:conclusions}

This work introduced Fed-IGTD, a horizontal FL for privacy-preserving tabular-to-image transformation based on IGTD. Fed-IGTD preserves the IGTD representation without requiring raw tabular data to leave individual clients. Results on five datasets demonstrated that the framework preserves utility under IID federation and showed that privacy protection applied only to the one-shot correlation-aggregation step incurs substantially lower utility costs than protecting the full training procedure with client-level DP-FedAvg, whose utility degradation becomes particularly pronounced in small-client regimes and under stringent privacy budgets. Fed-IGTD provides a principled framework for integrating tabular-to-image representation learning with horizontal federated learning and formal privacy guarantees, while offering a foundation for extending this approach to larger federations, heterogeneous feature schemas, secure aggregation, and broader privacy--utility analyses. The results demonstrate the potential of IGTD-based image representations to facilitate the application of deep learning to diverse tabular data domains within privacy-constrained and distributed settings.

\appendices
\section{Correlation matrices using centralized and federated approach}

\begin{figure*}
  \centering
  \begin{subfigure}[b]{0.85\linewidth}
    \includegraphics[width=\linewidth]{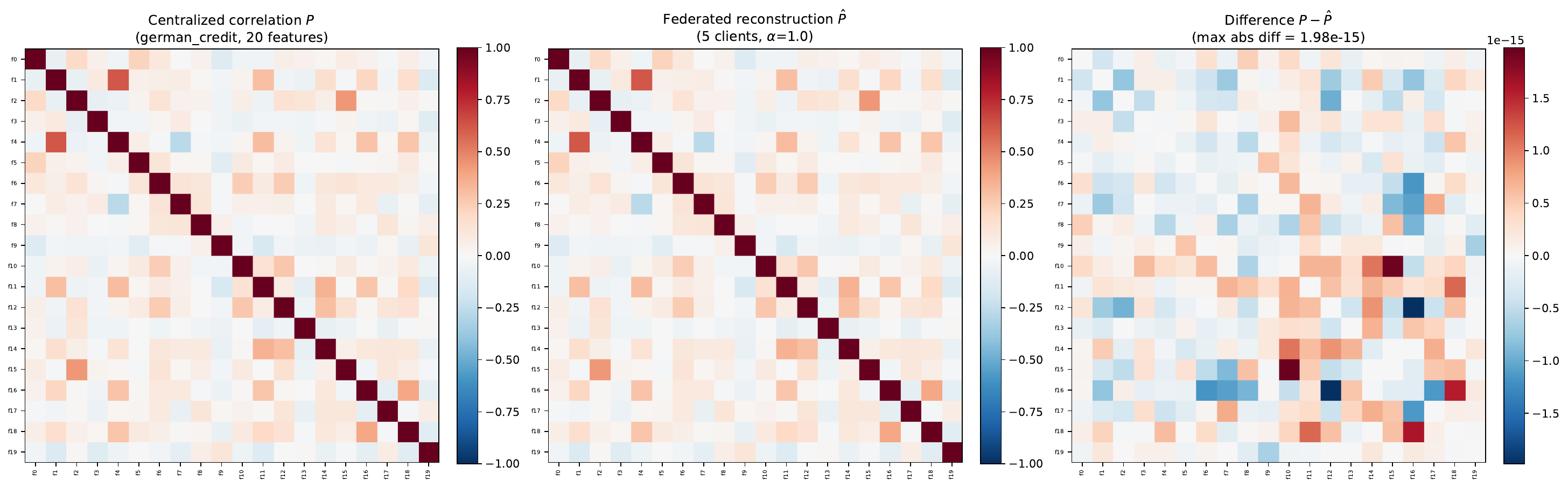} 
    \caption{} 
  \end{subfigure}
  \begin{subfigure}[b]{0.85\linewidth}
    \includegraphics[width=\linewidth]{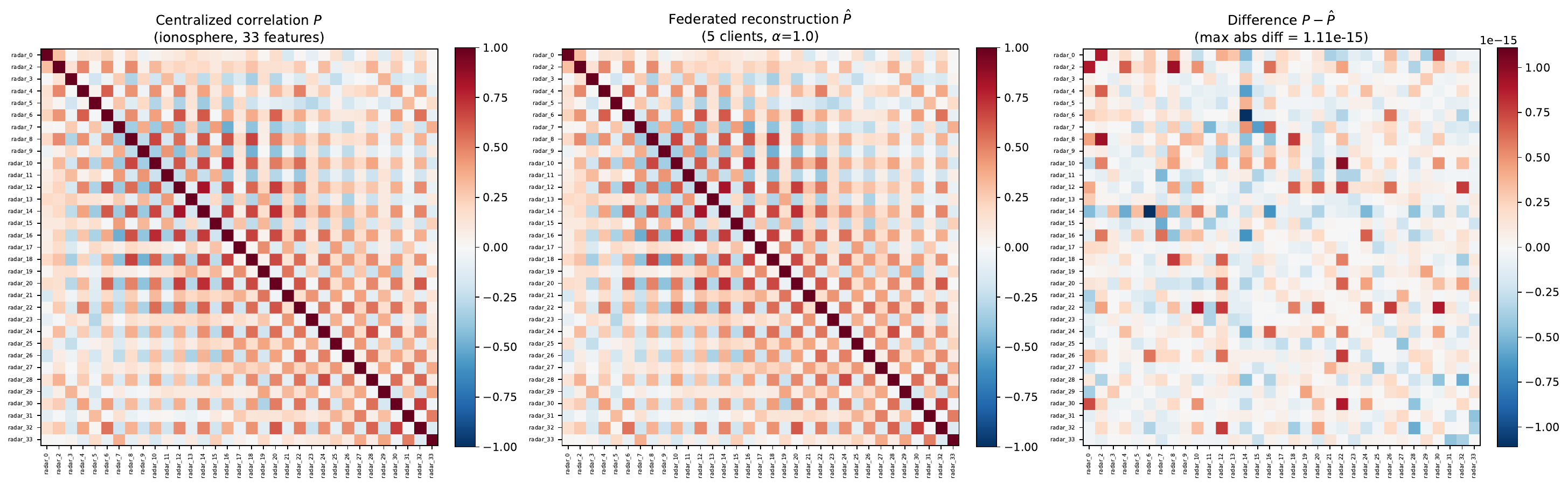} 
    \caption{} 
  \end{subfigure}
  \begin{subfigure}[b]{0.85\linewidth}
    \includegraphics[width=\linewidth]{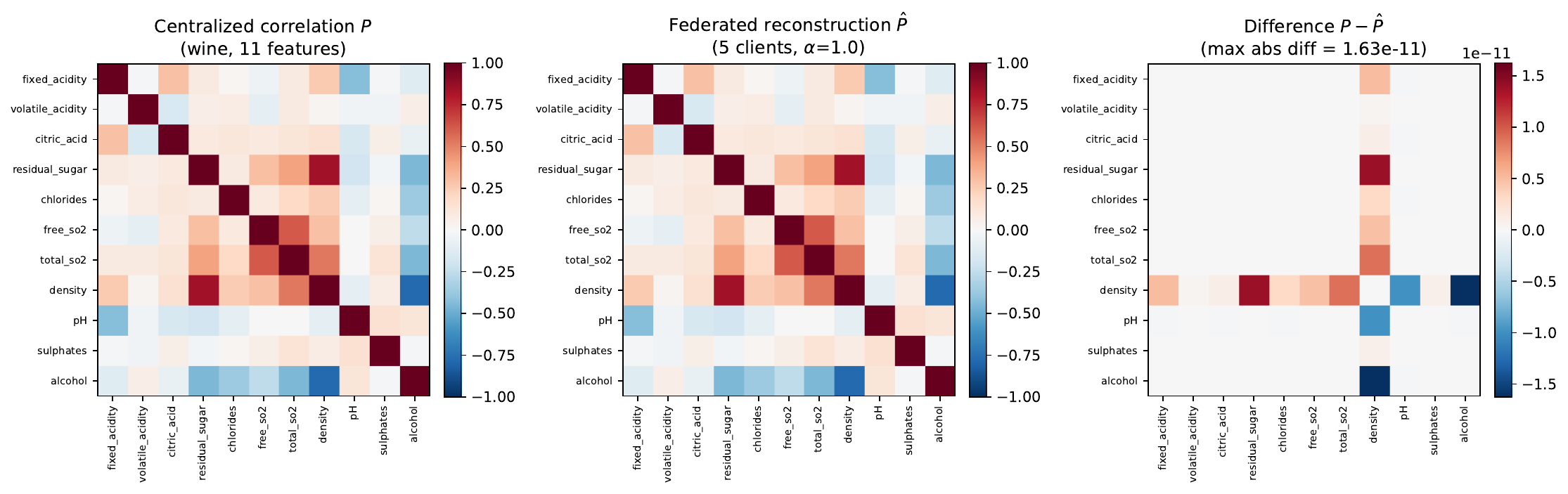} 
    \caption{} 
  \end{subfigure}
  \captionsetup{justification=justified, singlelinecheck=false, margin=6pt} 
  \caption{Comparison of centralized and federated correlation matrix. Considering the dataset of: (a) German Credit; (b) Ionosphere; and (c) Wine Quality.}
  \label{fig:heatmap_comparison_extension} 
\end{figure*}

\section{Dirichlet label-skew partitioning}

\begin{figure*}
  \centering
  \begin{subfigure}[b]{0.95\linewidth}
    \includegraphics[width=\linewidth]{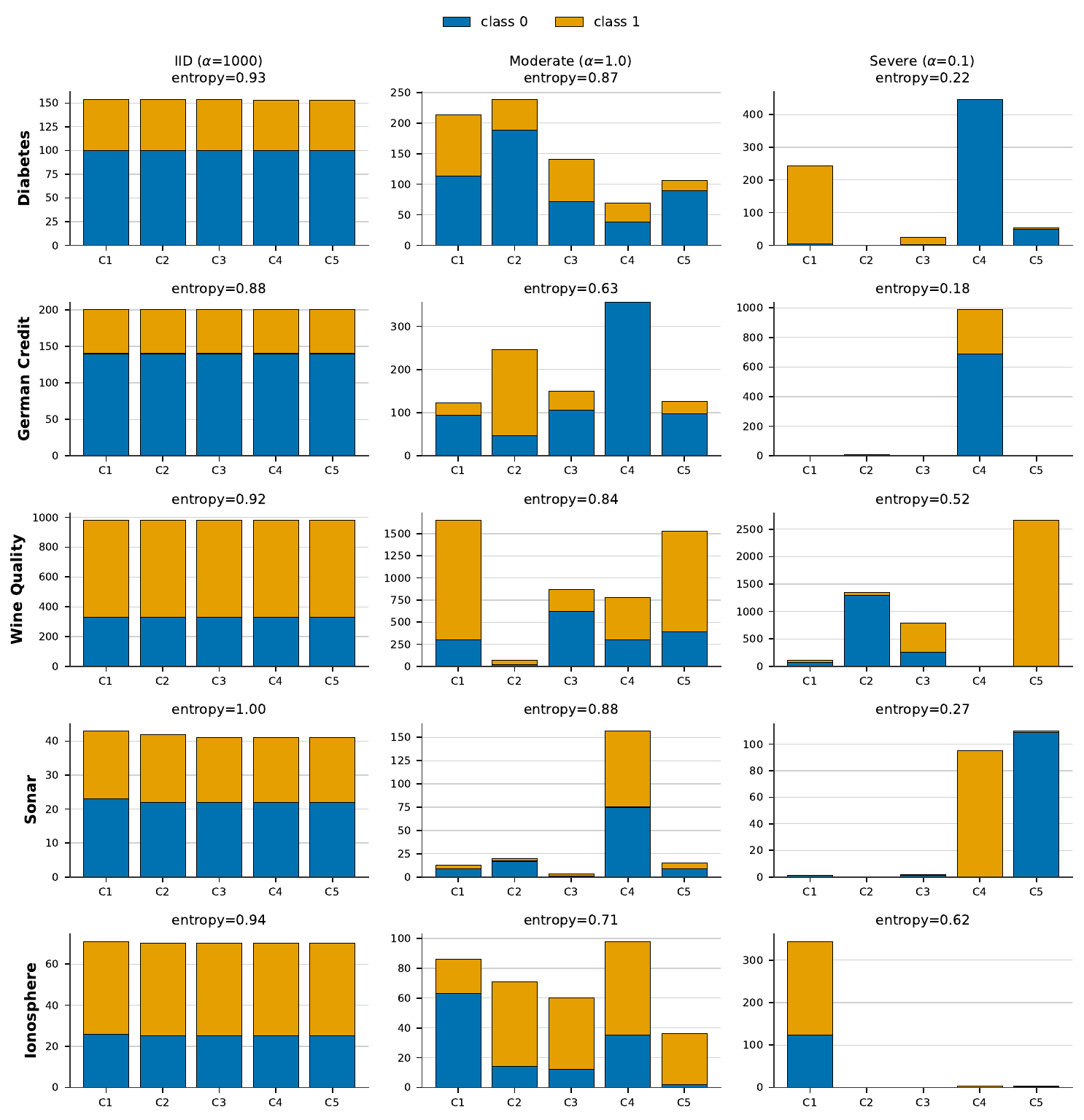} 
  \end{subfigure}
  \captionsetup{justification=justified, singlelinecheck=false, margin=6pt} 
  \caption{Dirichlet label-skew partitioning across non-IID severity for the datasets used in this study (5 clients each). Each row corresponds to one dataset; columns show the client-level sample counts and class balance under IID ($\alpha=1000$), Moderate ($\alpha=1.0$), and Severe ($\alpha=0.1$) partitioning. Bar height gives each client's total sample count, colored by class; the entropy annotation reports the mean per-client normalized class-balance entropy (1.0 = perfectly balanced across classes, 0.0 = single-class), averaged over all five clients.}
  \label{fig:dirichlet_supp} 
\end{figure*}


\ifCLASSOPTIONcompsoc
  \section*{Acknowledgments}
\else
  \section*{Acknowledgment}
\fi


\ifCLASSOPTIONcaptionsoff
  \newpage
\fi



%

\bibliographystyle{vancouver}
\bibliography{bare_adv_literature}

@article{zhu2021converting,
  title={Converting tabular data into images for deep learning with convolutional neural networks},
  author={Zhu, Yitan and Brettin, Thomas and Xia, Fangfang and Partin, Alexander and Shukla, Maulik and Yoo, Hyunseung and Evrard, Yvonne A and Doroshow, James H and Stevens, Rick L},
  journal={Scientific reports},
  volume={11},
  number={1},
  pages={11325},
  year={2021},
  publisher={Nature Publishing Group UK London}
}

@article{gomez2026tabular,
  title={Tabular-to-Image Encoding Methods for Melanoma Detection: A Proof-of-Concept},
  author={G{\'o}mez-Mart{\'\i}nez, Vanesa and Chushig-Muzo, David and Soguero-Ruiz, Cristina},
  journal={Applied Sciences},
  volume={16},
  number={5},
  pages={2459},
  year={2026},
  publisher={MDPI}
}

@article{lara2025transfer,
  title={Transfer learning for a tabular-to-image approach: A case study for cardiovascular disease prediction},
  author={Lara-Abelenda, Francisco J and Chushig-Muzo, David and Peiro-Corbacho, Pablo and G{\'o}mez-Mart{\'\i}nez, Vanesa and W{\"a}gner, Ana M and Granja, Concei{\c{c}}{\~a}o and Soguero-Ruiz, Cristina},
  journal={Journal of Biomedical Informatics},
  volume={165},
  pages={104821},
  year={2025},
  publisher={Elsevier}
}

@article{yuan2026fedvi,
  title={FedVI: Financial Cross-Domain Federated Learning with Scarce Overlapping Samples via Visual Representation of Heterogeneous Tabular Data and Meta-Optimization},
  author={Yuan, Kaiqing and Wu, Jiang},
  journal={Entropy},
  volume={28},
  number={6},
  pages={637},
  year={2026},
  publisher={MDPI}
}

@article{zhang2021survey,
  title={A survey on federated learning},
  author={Zhang, Chen and Xie, Yu and Bai, Hang and Yu, Bin and Li, Weihong and Gao, Yuan},
  journal={Knowledge-based systems},
  volume={216},
  pages={106775},
  year={2021},
  publisher={Elsevier}
}

@inproceedings{poonam2025hierarchical,
  title={Hierarchical classification of frontotemporal dementia subtypes utilizing tabular-to-image data conversion with deep learning methods},
  author={Poonam, Km and Kotra, Venkata Sathwik and Guha, Rajlakshmi and Chakrabarti, Partha P},
  booktitle={International Conference on Pattern Recognition},
  pages={386--401},
  year={2025},
  organization={Springer}
}

@article{wang2026diabetes,
  title={Diabetes Risk Modeling through Tabular-to-Image Transformations and Ensemble Learning},
  author={Wang, Shenghao and Upadhyay, Darshana and Zaman, Marzia and Naik, Kshirasagar and Rai, Shesh Nath},
  journal={Informatics in Medicine Unlocked},
  pages={101748},
  year={2026},
  publisher={Elsevier}
}

@article{yzzogh2025evaluating,
  title={Evaluating orb and sift with neural network as alternatives to cnn for traffic classification in sdn environments},
  author={Yzzogh, Hicham and Benaboud, Hafssa},
  journal={IEEe Access},
  volume={13},
  pages={51484--51498},
  year={2025},
  publisher={IEEE}
}

@article{sonmez2026comparative,
  title={Comparative evaluation of tabular-to-image deep learning pipelines for EEG-based epileptic seizure recognition: A comprehensive benchmark with statistical analysis and explainable AI},
  author={S{\"o}nmez, Mesut Ersin},
  journal={Computers in Biology and Medicine},
  volume={214},
  pages={111896},
  year={2026},
  publisher={Elsevier}
}

@article{hosseini2023knock,
  title={Knock probability determination employing convolutional neural network and IGTD algorithm},
  author={Hosseini, M and Chitsaz, I},
  journal={Energy},
  volume={284},
  pages={129282},
  year={2023},
  publisher={Elsevier}
}

@inproceedings{nazarri2023generating,
  title={Generating network intrusion image through IGTD algorithm for CNN classification},
  author={Nazarri, Muhammad Nur Aman Ahmad and Yusof, Mohammad Hafiz Mohd and Almohammedi, Akram A},
  booktitle={2023 3rd International Conference on Computing and Information Technology (ICCIT)},
  pages={172--177},
  year={2023},
  organization={IEEE}
}

@article{li2020federated,
  title={Federated optimization in heterogeneous networks},
  author={Li, Tian and Sahu, Anit Kumar and Zaheer, Manzil and Sanjabi, Maziar and Talwalkar, Ameet and Smith, Virginia},
  journal={Proceedings of Machine learning and systems},
  volume={2},
  pages={429--450},
  year={2020}
}

@article{li2021fedbn,
  title={Fedbn: Federated learning on non-iid features via local batch normalization},
  author={Li, Xiaoxiao and Jiang, Meirui and Zhang, Xiaofei and Kamp, Michael and Dou, Qi},
  journal={arXiv preprint arXiv:2102.07623},
  year={2021}
}

@article{mcmahan2017learning,
  title={Learning differentially private recurrent language models},
  author={McMahan, H Brendan and Ramage, Daniel and Talwar, Kunal and Zhang, Li},
  journal={arXiv preprint arXiv:1710.06963},
  year={2017}
}

@inproceedings{mironov2017renyi,
  title={R{\'e}nyi differential privacy},
  author={Mironov, Ilya},
  booktitle={2017 IEEE 30th computer security foundations symposium (CSF)},
  pages={263--275},
  year={2017},
  organization={IEEE}
}

@article{chua2024scalable,
  title={Scalable DP-SGD: Shuffling vs. poisson subsampling},
  author={Chua, Lynn and Ghazi, Badih and Kamath, Pritish and Kumar, Ravi and Manurangsi, Pasin and Sinha, Amer and Zhang, Chiyuan},
  journal={Advances in Neural Information Processing Systems},
  volume={37},
  pages={70026--70047},
  year={2024}
}

@inproceedings{mcmahan2017communication,
  title={Communication-efficient learning of deep networks from decentralized data},
  author={McMahan, Brendan and Moore, Eider and Ramage, Daniel and Hampson, Seth and y Arcas, Blaise Aguera},
  booktitle={Artificial intelligence and statistics},
  pages={1273--1282},
  year={2017},
  organization={Pmlr}
}

@article{chen2024federated,
  title={When federated learning meets privacy-preserving computation},
  author={Chen, Jingxue and Yan, Hang and Liu, Zhiyuan and Zhang, Min and Xiong, Hu and Yu, Shui},
  journal={ACM Computing Surveys},
  volume={56},
  number={12},
  pages={1--36},
  year={2024},
  publisher={ACM New York, NY}
}

@article{bazgir2020representation,
  title={Representation of features as images with neighborhood dependencies for compatibility with convolutional neural networks},
  author={Bazgir, Omid and Zhang, Ruibo and Dhruba, Saugato Rahman and Rahman, Raziur and Ghosh, Souparno and Pal, Ranadip},
  journal={Nature communications},
  volume={11},
  number={1},
  pages={4391},
  year={2020},
  publisher={Nature Publishing Group UK London}
}

@article{sharma2019deepinsight,
  title={DeepInsight: A methodology to transform a non-image data to an image for convolution neural network architecture},
  author={Sharma, Alok and Vans, Edwin and Shigemizu, Daichi and Boroevich, Keith A and Tsunoda, Tatsuhiko},
  journal={Scientific reports},
  volume={9},
  number={1},
  pages={11399},
  year={2019},
  publisher={Nature Publishing Group UK London}
}

@article{castillo2023tinto,
  title={TINTO: converting tidy data into image for classification with 2-dimensional convolutional neural networks},
  author={Castillo-Cara, Manuel and Talla-Chumpitaz, Reewos and Garc{\'\i}a-Castro, Ra{\'u}l and Orozco-Barbosa, Luis},
  journal={SoftwareX},
  volume={22},
  pages={101391},
  year={2023},
  publisher={Elsevier}
}

@article{jiang2026representation,
  title={Representation learning for tabular data: A comprehensive survey},
  author={Jiang, Jun-Peng and Liu, Si-Yang and Cai, Hao-Run and Zhou, Qi-Le and Ye, Han-Jia},
  journal={IEEE Transactions on Pattern Analysis and Machine Intelligence},
  year={2026},
  publisher={IEEE}
}

@article{somvanshi2026survey,
  title={A survey on tabular data: from tree-based methods to tabular deep learning},
  author={Somvanshi, Shriyank and Das, Subasish and Javed, Syed and Antariksa, Gian and Hossain, Ahmed},
  journal={ACM Computing Surveys},
  year={2026},
  publisher={ACM New York, NY}
}

@article{wang2026robust,
  title={Robust and Interpretable Tabular Data Classification via Multi-Channel Image Conversion and Channel-Wise Gating},
  author={Wang, Zengshuai and Zheng, Minhua and Liu, Peter Xiaoping},
  journal={Neural Networks},
  pages={109316},
  year={2026},
  publisher={Elsevier}
}

@article{shenoy2025exploring,
  title={Exploring privacy mechanisms and metrics in federated learning},
  author={Shenoy, Dhanya and Bhat, Radhakrishna and Krishna Prakasha, K},
  journal={Artificial Intelligence Review},
  volume={58},
  number={8},
  pages={223},
  year={2025},
  publisher={Springer}
}

@article{liu2025comprehensive,
  title={A comprehensive benchmark of spatial encoding methods for tabular data with deep neural networks},
  author={Liu, Jiayun and Castillo-Cara, Manuel and Garc{\'\i}a-Castro, Ra{\'u}l},
  journal={Information Fusion},
  pages={104088},
  year={2025},
  publisher={Elsevier}
}

@article{chushig2026tabsom,
  title={TabSOM: A tabular-to-image encoding method based on self-organizing maps},
  author={Chushig-Muzo, David and de Cara, Mar{\'\i}a {\'A}ngeles Rodr{\'\i}guez and Milara, Eva and Lara-Abelenda, Francisco J and Zhinin-Vera, Luis and Peluffo-Ord{\'o}{\~n}ez, Diego H},
  journal={arXiv preprint arXiv:2608.13513},
  year={2026}
}



%








\end{document}